\documentclass[preprint,12pt,authoryear]{elsarticle}
\usepackage{hyperref}
\hypersetup{pdftitle=GASwarm,pdfauthor=Puente,pdfsubject=Puente,pdfkeywords={Puente},pdfcreator={Puente}}

\usepackage{graphicx}
\usepackage{subfigure}
\usepackage{amsmath}
\usepackage{multicol}
\usepackage{multirow}
\usepackage{url}
\usepackage[capitalize]{cleveref}
\usepackage[table,xcdraw]{xcolor}
\usepackage{tabularx}
\newcolumntype{C}[1]{>{\centering\arraybackslash}p{#1}}
\usepackage{comment}

\usepackage[ruled,vlined]{algorithm2e}

\graphicspath{ {./} }

\usepackage{amssymb}
\journal{arXiv}

\begin{document}

\begin{frontmatter}



%\begin{titlepage}
%\begin{center}
%\vspace*{1cm}

%\textbf{Evolutionary-Based System for Path Planning for Unmanned Aerial Vehicles Swarms in Obstacle Environments}

%\vspace{1.5cm}

% Author names and affiliations
%Alejandro Puente-Castro$^{a}$ (a.puentec@udc.es), Enrique Fernandez-Blanco$^a$ (enrique.fernandez@udc.es), Daniel Rivero$^a$ (daniel.rivero@udc.es)\\

%\hspace{10pt}

%\begin{flushleft}
%\small  
%$^a$ Faculty of Computer Science, CITIC, University of A Coruna, A Coruna, 15007, Spain \\

%\vspace{1cm}
%\textbf{Corresponding Author:} \\
%Last Author \\
%Full address of the corresponding author, including the country name \\
%Tel: (555) 555-1234 \\
%Email: last.author@mail.com

%\end{flushleft}        
%\end{center}
%\end{titlepage}

\title{Genetic Algorithm Based System for Path Planning with Unmanned Aerial Vehicles Swarms in Cell-Grid Environments}


\author[udc]{Alejandro Puente-Castro}
\ead{a.puentec@udc.es}

\author[udc]{Enrique Fernandez-Blanco}

\author[udc]{Daniel Rivero}

\affiliation[udc]{organization={Faculty of Computer Science, CITIC},
             addressline={University of A Coruna},
             city={A Coruna},
             postcode={15007},
             country={Spain}}

\begin{abstract}
Path Planning methods for autonomously controlling swarms of unmanned aerial vehicles (UAVs) are gaining momentum due to their operational advantages. An increasing number of scenarios now require autonomous control of multiple UAVs, as autonomous operation can significantly reduce labor costs. Additionally, obtaining optimal flight paths can lower energy consumption, thereby extending battery life for other critical operations. Many of these scenarios, however, involve obstacles such as power lines and trees, which complicate Path Planning. This paper presents an evolutionary computation-based system employing genetic algorithms to address this problem in environments with obstacles. The proposed approach aims to ensure complete coverage of areas with fixed obstacles, such as in field exploration tasks, while minimizing flight time—regardless of map size or the number of UAVs in the swarm. No specific goal points or prior information beyond the provided map is required. The experiments conducted in this study used five maps of varying sizes and obstacle densities, as well as a control map without obstacles, with different numbers of UAVs. The results demonstrate that this method can determine optimal paths for all UAVs during full map traversal, thus minimizing resource consumption. A comparative analysis with other state-of-the-art approach is presented to highlight the advantages and potential limitations of the proposed method.

\end{abstract}



\begin{keyword}

UAV \sep Swarm \sep Evolutionary Computation \sep Genetic Algorithm \sep Obstacle

\end{keyword}

\end{frontmatter}


\section{Introduction}\label{introduction}

A variety of industrial and emergency problems are being solved using swarms of unmanned aerial vehicles (UAVs), as highlighted in recent studies \citep{albani2017monitoring, huuskonen2018soil, CORTE2020105815, bocchino2018f, rabinovitch2021scaling, liu2018motif}. UAVs have gained popularity due to several factors, including their cost-effectiveness, excellent mobility, operational safety, and compact size, which is particularly advantageous for specific maneuvers \citep{yeaman1998virtual}. Furthermore, the wide range of UAV models available on the market enhances their versatility, as they can be equipped with various types of sensors tailored to diverse applications. In particular, sensor development for UAVs has become increasingly significant, especially in the field of remote sensing \citep{noor2018remote}. Thanks to their flexible configurations—such as adaptable architectures and sensor payloads—UAVs are now widely employed in numerous fields \citep{austin2011unmanned}.

One major challenge faced by UAVs is their high power consumption, which limits their operational time. Due to their compact size, finding lightweight, high-capacity power sources remains a complex task. However, equipping UAVs with efficient and lightweight power solutions can significantly extend their flight duration while maintaining their operational capabilities. This improvement enables longer missions and enhances maneuverability.

When deploying groups or swarms of UAVs, battery limitations can be mitigated. By operating multiple UAVs simultaneously, tasks can be completed more efficiently, as shorter flight paths are required. This reduces the likelihood of individual units depleting their batteries mid-operation. Consequently, energy consumption per UAV decreases, minimizing the risk of mission interruptions or potential failures.

UAV swarms, like other types of robotic swarms, hold vast potential for real-world applications. The primary advantage of swarm robotics is its inherent robustness, which can manifest in various ways. Firstly, a swarm consists of numerous simple agents that are not assigned specific roles or responsibilities, enabling self-organization and dynamic reorganization of individual robots' deployment. Furthermore, the swarm approach is highly resilient to individual agent failures due to its distributed control; there is no single point of failure or vulnerability. This high level of robustness seen in UAV swarms is an inherent feature of swarm robotics, unlike the significant technical investment required for fault tolerance in conventional robotic systems \citep{sahin2008special}.

Initial flying tests with swarms required one UAV operator per UAV, which significantly increased operational costs when deployed in groups. However, recent advances in algorithms and communication technologies have revolutionized swarm control. With these developments, a single operator can efficiently manage the entire swarm, reducing operational expenses \citep{zhao2018survey, campion2018review}. These improvements facilitate faster and more effective communication between UAVs and enable more precise collision avoidance path calculations. Consequently, human intervention in potentially dangerous situations is no longer necessary.

Consequently, recent methods have been aiming to provide the entire swarm complete autonomous control. These paths are carefully designed to maximize effectiveness while minimizing expenses. This challenge, commonly referred to as the Path Planning Problem \citep{aggarwal2020path}, focuses on mapping the trajectories of autonomous devices like UAVs. Since many UAV operations are conducted at low altitudes, these aircraft must maneuver around obstructions in their operational area. When calculating flight paths and forecasting the future locations of each UAV in the swarm, it is essential to take these obstacles into account in order to guarantee collision-free operations throughout the fleet. Therefore, the goal is to design routes that are both effective and able to steer clear of obstacles and other UAVs as much as feasible.

The Swarm Intelligence (SI) paradigm provides a variety of techniques to address the complexity of this type of development \citep{kennedy2006swarm}. These algorithms are designed to coordinate large numbers of agents simultaneously. This coordination relies on a collection of independent actors operating in a powerful, self-organized manner while adhering to fundamental rules \citep{bonabeau2001swarm, beckerleg2016evolving}. In other words, each UAV in the swarm acts as a unique agent. Each agent possesses its own information and behavior, which are influenced by the system's rules and the information exchanged with other agents. This coordinated behavior is ultimately aimed at achieving a common objective in the most efficient way possible \citep{stentz1997optimal}.

In military applications, some of these Path Planning algorithms have already been implemented. In contrast, civilian applications remain limited and are primarily employed for tasks such as navigating urban environments \citep{puente2021review}. Despite the vast potential of these technologies, there is a notable lack of solutions specifically designed for agricultural and forestry applications, particularly those aimed at optimizing field prospecting tasks.

The objective of this study is to develop a system that employs Evolutionary Computation techniques to address the Path Planning problem in 2D grid-based maps with fixed obstacles and variable numbers of UAVs. The primary contributions of this work are as follows:

\begin{enumerate}
    \item A novel Evolutionary Computation-based system capable of determining the optimal flight path for a UAV swarm to achieve complete map coverage, minimizing flight time and, consequently, reducing resource consumption such as battery usage.
    \item A system that can predict the flight path for any number of UAVs on maps of varying sizes and obstacle types.
    \item A comparative analysis of maps with obstacles versus maps without obstacles.
    \item A comprehensive performance comparison between the proposed Evolutionary Computation-based system and a state-of-the-art Path Planning algorithm.
\end{enumerate}

This paper is organized as follows: Section~\ref{background} presents a summary of the most relevant state-of-the-art publications; Section~\ref{materials} details the technical requirements for the development of the proposed method; Section~\ref{results} summarizes the results obtained from the experimental process; Section~\ref{conclusions} provides the conclusions drawn from analyzing the results; and, finally, Section~\ref{future_work} outlines potential future works and studies that could address the problem at hand.

\section{Background}\label{background}

In the current state-of-the-art, two main research categories are prominent: Reinforcement Learning (RL) \citep{sutton2018reinforcement} and Evolutionary Computation (EC) \citep{holland1992genetic}, as highlighted by \citet{puente2021review}. Each approach offers distinct methodologies for addressing the Path Planning Problem, particularly in complex environments such as those involving UAV swarms.

An example of Reinforcement Learning is presented by \citet{qiu2022data}, who employ an Actor-Critic RL system to manage multiple UAVs simultaneously, where each UAV relies solely on locally available environmental data. This Actor-Critic approach is also applied by \citet{wei2022high} in their study on large-scale data collection, proposing a time-based energy consumption model. However, this method does not account for movement types, potentially resulting in higher energy usage in paths with frequent directional changes. Additionally, \citet{qie2019joint} introduced the Multi-Agent Deep Deterministic Policy Gradient (MADDPG) system to handle target assignment and trajectory planning for multiple UAVs, achieving real-time performance. In the field of communication, \citet{peng2019anti} proposed a multi-parameter RL system to enhance UAV swarm communications. More recently, \citet{lee2022federated} presented a federated RL-based system that integrates federated learning with RL for aerial remote sensing tasks, providing a decentralized approach to enhance UAV sensing capabilities. Similarly, \citet{chen2022multi} proposed a multi-agent RL-based algorithm for autonomous Path Planning in reconnaissance missions under incomplete information, using a centralized training and decentralized execution strategy with a joint reward function that balances time cost, coverage, and security.

The Evolutionary Computation (EC) branch, particularly Genetic Algorithms (GA), has also been extensively applied in robotics and UAV control. For instance, \citet{zhang2023path} introduced a GA to navigate robots on grid-based maps, encoding paths as genetic sequences. A limitation of this approach is that paths may pass through grid intersection points, complicating cell identification. Similarly, \citet{wu2023field} presented a GA-based system for transplanting robots to cover an entire area within a grid map, defining the layout and obstacles without specifying start or goal points, though the system offers limited movement variety. GA has also been applied to vertical maps, as demonstrated by \citet{guban2022path} in modeling warehouse shelves—a division commonly used in vertical operations due to its ease of regional data addition \citep{tao2022path}. In a 3D environment, \citet{kong2022improved} proposed a GA for swarm control, using a simulator to validate their approach, which avoids local maxima but may incur high computational costs. Similarly, \citet{liu2022improved} suggested a GA for handling complex terrains in 3D, providing smoothed paths without requiring post-processing, with the distance to target points serving as the fitness function.

In 2020, \citet{toorchi2020skeleton} introduced the SSR protocol, an RL-based swarm control technique designed to generate smoother paths by leveraging the structural configuration of the swarm. Likewise, \citet{xiang2020research} proposed an RL-based system for defending UAV swarms against potential threats by dynamically adjusting flight speed and angles in response to attacks. For mission-oriented operations, \citet{park2020vmcs} developed the Versatile Multi-Vehicle Control System (VMCS), an Artificial Potential Field (APF)-based system aimed at enhancing mission efficiency through adaptive topology control. More recently, \citet{luo2021grpavoid} proposed a group-level EC mechanism focused on collision avoidance, utilizing a multi-level model with adaptive fitness assessment. Similarly, \citet{stolfi2021uav} presented a Competitive Coevolutionary Genetic Algorithm (CompCGA) to optimize swarm surveillance and evasion capabilities, demonstrating superior performance compared to existing methods.

A small number of maps with predetermined sizes are used to test the majority of the listed works. In other words, they test their algorithms in a static obstacle topology without experimenting with various sizes. A CE-based system will be tested on various maps with various obstacle topologies, according to the proposed article.

\section{Materials and Method}\label{materials}

\subsection{Problem Formulation}\label{sec:formulation}

The primary objective of this study is to develop a system that addresses the Path Planning Problem \cite{aggarwal2020path} for UAV swarms in environments with obstacles. Effective Path Planning in scenarios involving multiple UAVs requires the consideration of various factors to ensure efficiency, control, robust collaboration, and safety. Addressing these challenges is critical to achieving the study's objectives.

Path Planning is conceptualized here as a combination of interrelated problems, a perspective widely recognized in the literature \citep{puente2022uav}. Consequently, the Path Planning problem is structured into the following key areas:

\begin{itemize}
    \item Definition of Flight Maps
    \item UAV Movement Dynamics
    \item Design of the Proposed Model
    \item Model Optimization
    \item Evaluation Metrics for Model Performance
\end{itemize}

A central concept in addressing the Path Planning problem involves simplifying environmental, movement, and other variable factors \citep{giesbrecht2004global}. In real-world applications, UAVs navigate through complex, continuous environments with infinitely many points, requiring the exploration of vast path combinations to determine the optimal path. To manage this complexity, a cell-based map representation is commonly applied. By dividing the map into finite cells, the exploration process becomes more manageable. However, this representation may omit critical terrain details, complicating optimization metrics such as path length and often oversimplifying real-world environments.

The movement capabilities of UAVs are also taken into account. While UAVs exhibit high stability during flight, their movements are inherently complex, often resulting from combinations of simpler maneuvers \citep{susanto2021application}. To streamline path calculations and enhance coverage, UAV movements are modeled as discrete actions, ignoring curves or altitude changes. This simplification allows for straightforward assessments of whether a UAV can reach a specific cell. However, it may result in abrupt path changes, where smoother curves could produce more efficient trajectories.

A key challenge of this approach is its tendency to converge on local optima \citep{jaakkola1994reinforcement}. This convergence may lead to satisfactory but suboptimal solutions, emphasizing the need for alternative methods to mitigate the risk of such outcomes. To evaluate the effectiveness of this approach, a diverse set of maps previously published in a peer-reviewed study within the state-of-the-art is used \citep{puente2024q}.

\subsection{Genetic Algorithms}\label{sec:GeneticAlgorithms}

The calculation of flight paths is performed using Genetic Algorithms (GAs) \citep{holland1984genetic}. These algorithms simulate a population of potential solutions, where each solution evolves through selection, crossover, and mutation, mimicking the process of natural selection. The most adapted individuals represent the best solutions to the problem.

In a GA, each solution is encoded as an individual in a population, with the solution represented by a genotype—typically a string of binary or real values. An initial population is generated, and a crossover rate is defined, determining the probability that two individuals will combine their genotypes to produce offspring. Parent selection for crossover is guided by a satisfaction criterion, ensuring controlled population growth and diversity.

To introduce variability, a mutation rate is applied, defining the probability that specific genes within an individual's chromosome will mutate. This variability promotes beneficial changes and accelerates the algorithm's convergence towards an optimal solution.

\subsection{Maps}\label{sec:Maps}

To evaluate the proposed approach, six different maps with varying topologies were used, selected to reflect diverse scenarios and to highlight potential limitations of the method. The chosen topologies include various obstacle arrangements to test the algorithm’s robustness and adaptability, as detailed in Sections~\ref{sec:Limitations} and the results section.

Each map is represented as a grid of cells, allowing UAVs to perform movements between contiguous cells. This grid-based representation simplifies Path Planning by discretizing the continuous environment into manageable units. Each UAV has four possible discrete movements: up, down, left, and right. This simplification ignores complex maneuvers, such as diagonal movements or altitude changes, focusing solely on horizontal and vertical coverage in order to get a more precise metri for evaluating the results. 

\begin{figure}[!ht]
    \centering
    \subfigure[5$\times$5]{\includegraphics[scale=0.4]{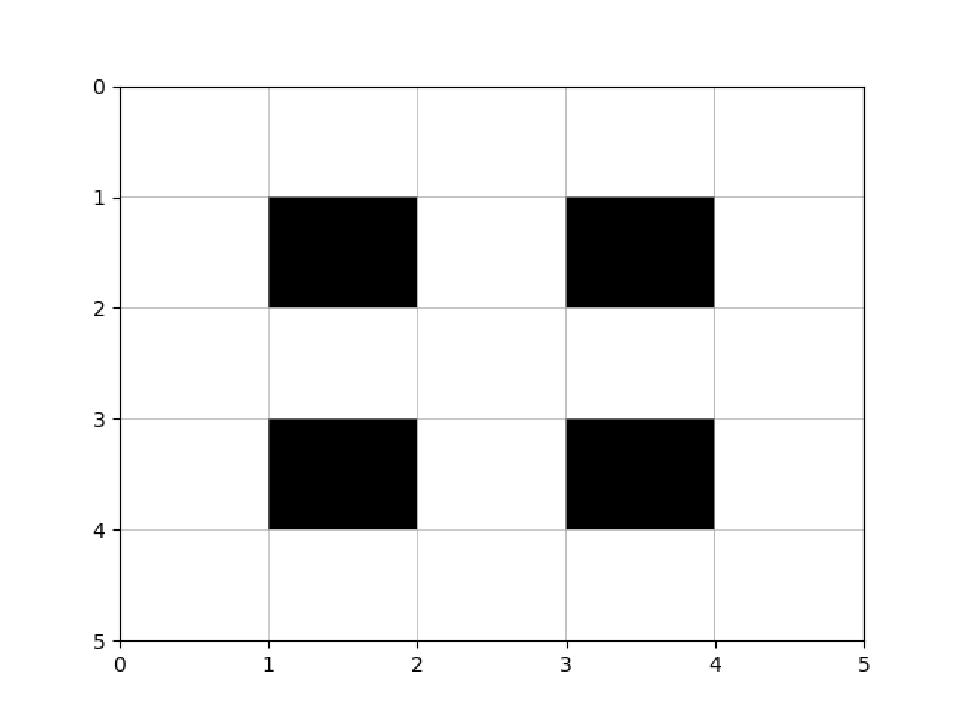}\label{fig:map_set1}}
    \subfigure[6$\times$6]{\includegraphics[scale=0.4]{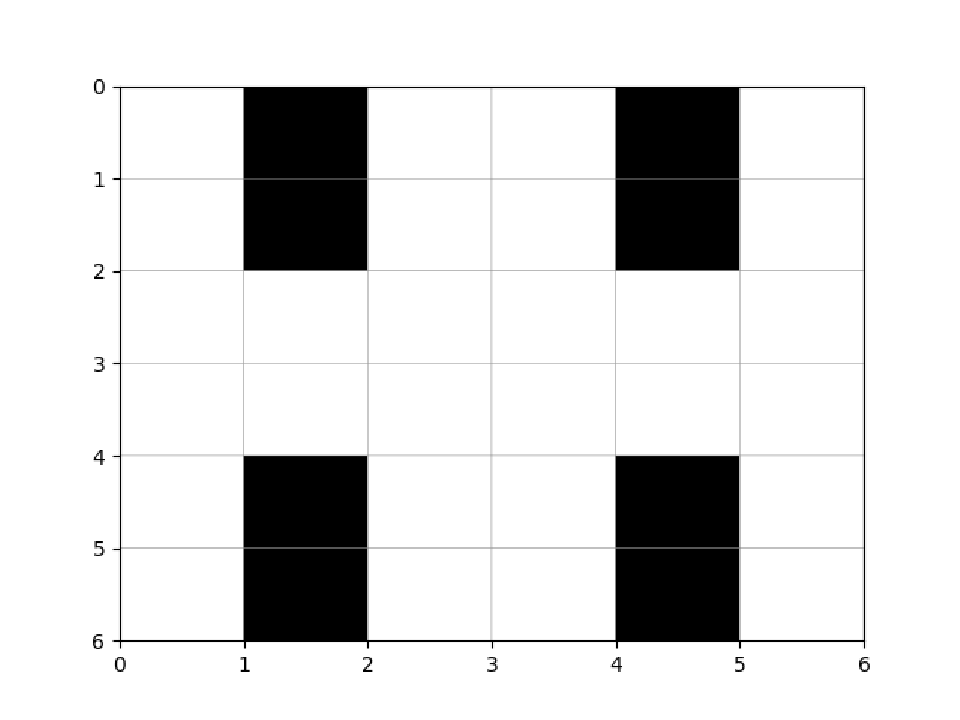}\label{fig:map_set2}}
    \subfigure[7$\times$7]{\includegraphics[scale=0.38]{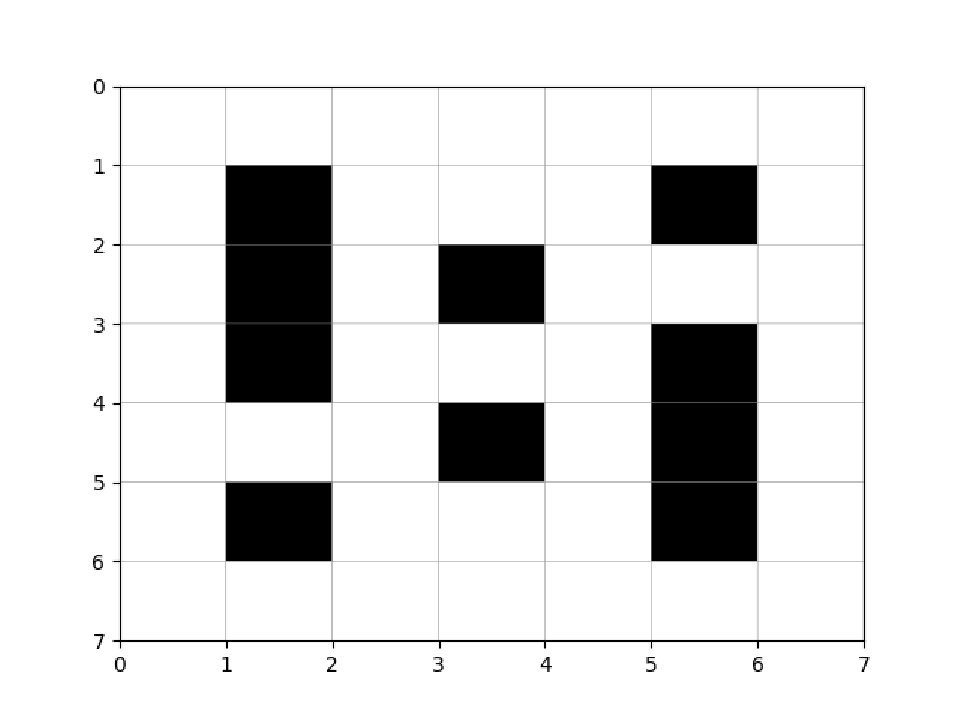}\label{fig:map_set3}}
    \subfigure[8$\times$8]{\includegraphics[scale=0.38]{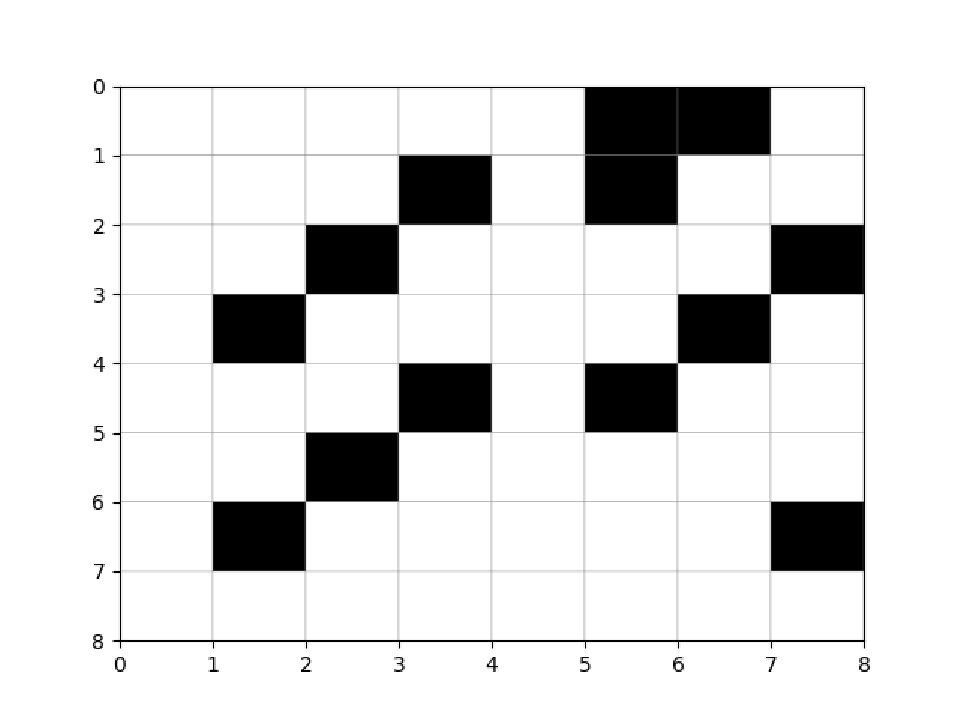}\label{fig:map_set4}}
    \subfigure[9$\times$9]{\includegraphics[scale=0.38]{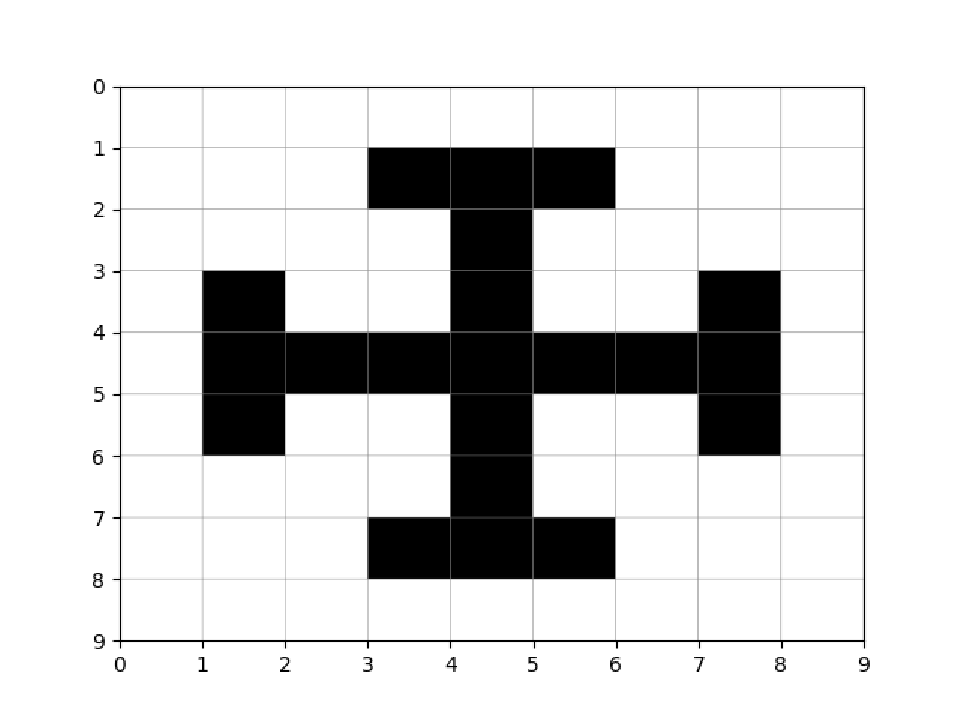}\label{fig:map_set5}}
    \caption{The flight environment maps. Black cells represent obstacles, while white cells indicate areas that UAVs in the swarm must cover. The diversity in map designs allows testing under varying complexity levels.}
    \label{fig:map_set}
\end{figure}

The map set was designed with increasing complexity to evaluate the system's scalability and robustness \citep{puente2024q}:
\begin{itemize}
    \item \textbf{Map 1:} A 7x7 cell grid without obstacles, representing the simplest scenario for evaluating the baseline performance of the algorithm.
    \item \textbf{Map 2 (Figure~\ref{fig:map_set1}):} A smaller map with central obstacles and 21 visitable cells, requiring UAVs to travel previously traversed paths, testing the algorithm's ability to handle revisits.
    \item \textbf{Map 3 (Figure~\ref{fig:map_set2}):} Contains multiple corners, increasing the difficulty of coverage planning, especially when fewer UAVs are deployed. It has 28 visitable cells.
    \item \textbf{Map 4 (Figure~\ref{fig:map_set3}):} Requires revisiting cells to achieve full coverage of 39 visitable cells, representing irregular topologies common in real-world scenarios.
    \item \textbf{Map 5 (Figure~\ref{fig:map_set4}):} A highly irregular map with numerous corners and 51 visitable cells, introducing significant challenges in coverage optimization.
    \item \textbf{Map 6 (Figure~\ref{fig:map_set5}):} Features a large central obstacle and 60 visitable cells. This map exemplifies scenarios where collaborative UAV operations are essential.
\end{itemize}

For each map, experiments involved deploying between 1 and 4 UAVs, with initial positions at the four corners: top-left, bottom-left, top-right, and bottom-right. The goal was to evaluate how UAV team size influences coverage efficiency and whether the proposed system scales effectively with increasing numbers of UAVs.

% Table~\ref{tab:NumUnvisitedCells} summarizes the number of cells each UAV must visit for complete coverage in each map. The data highlight the challenge posed by increasing obstacle density and map size.

%\begin{table}[]
%\centering
%\caption{Number of cells to visit for each map (rows) and each number of UAVs (columns).}
%\begin{tabular}{cc|cccc|}
%\cline{3-6}
%                                           &   & \multicolumn{4}{c|}{Number of UAVs}                                               \\ \cline{3-6} 
%                                           &   & \multicolumn{1}{c|}{1}  & \multicolumn{1}{c|}{2}  & \multicolumn{1}{c|}{3}  & 4  \\ \hline
%\multicolumn{1}{|c|}{\multirow{6}{*}{Map}} & 1 & \multicolumn{1}{c|}{48} & \multicolumn{1}{c|}{47} & \multicolumn{1}{c|}{46} & 45 \\ \cline{2-6} 
%\multicolumn{1}{|c|}{}                     & 2 & \multicolumn{1}{c|}{20} & \multicolumn{1}{c|}{19} & \multicolumn{1}{c|}{18} & 17 \\ \cline{2-6} 
%\multicolumn{1}{|c|}{}                     & 3 & \multicolumn{1}{c|}{27} & \multicolumn{1}{c|}{26} & \multicolumn{1}{c|}{25} & 24 \\ \cline{2-6} 
%\multicolumn{1}{|c|}{}                     & 4 & \multicolumn{1}{c|}{38} & \multicolumn{1}{c|}{37} & \multicolumn{1}{c|}{36} & 35 \\ \cline{2-6} 
%\multicolumn{1}{|c|}{}                     & 5 & \multicolumn{1}{c|}{50} & \multicolumn{1}{c|}{49} & \mulicolumn{1}{c|}{48} & 47 \\ \cline{2-6} 
%\multicolumn{1}{|c|}{}                     & 6 & \multicolumn{1}{c|}{59} & \multicolumn{1}{c|}{58} & \multicolumn{1}{c|}{57} & 56 \\ \hline
%\end{tabular}
%\label{tab:NumUnvisitedCells}
%\end{table}

The optimal number of movements required to cover each map was calculated and is presented in Table~\ref{tab:OptimalNumMovements}. Given that obstacles increase complexity, the actual maximum number of epochs was set to twice the theoretical minimum, providing a realistic benchmark for performance evaluation.

\begin{table}[]
\centering
\caption{Optimal number of movements for each map (rows) and each number of UAVs (columns).}
\begin{tabular}{cc|cccc|}
\cline{3-6}
                                           &   & \multicolumn{4}{c|}{Number of UAVs}                                               \\ \cline{3-6} 
                                           &   & \multicolumn{1}{c|}{1}  & \multicolumn{1}{c|}{2}  & \multicolumn{1}{c|}{3}  & 4  \\ \hline
\multicolumn{1}{|c|}{\multirow{6}{*}{Map}} & 1 & \multicolumn{1}{c|}{48} & \multicolumn{1}{c|}{24} & \multicolumn{1}{c|}{16} & 12 \\ \cline{2-6} 
\multicolumn{1}{|c|}{}                     & 2 & \multicolumn{1}{c|}{20} & \multicolumn{1}{c|}{10} & \multicolumn{1}{c|}{6}  & 5  \\ \cline{2-6} 
\multicolumn{1}{|c|}{}                     & 3 & \multicolumn{1}{c|}{27} & \multicolumn{1}{c|}{13} & \multicolumn{1}{c|}{9}  & 6  \\ \cline{2-6} 
\multicolumn{1}{|c|}{}                     & 4 & \multicolumn{1}{c|}{38} & \multicolumn{1}{c|}{19} & \multicolumn{1}{c|}{12} & 9  \\ \cline{2-6} 
\multicolumn{1}{|c|}{}                     & 5 & \multicolumn{1}{c|}{50} & \multicolumn{1}{c|}{25} & \multicolumn{1}{c|}{16} & 12 \\ \cline{2-6} 
\multicolumn{1}{|c|}{}                     & 6 & \multicolumn{1}{c|}{59} & \multicolumn{1}{c|}{29} & \multicolumn{1}{c|}{19} & 14 \\ \hline
\end{tabular}
\label{tab:OptimalNumMovements}
\end{table}

\subsection{Model}\label{sec:Model}

The proposed method employs a Genetic Algorithm (GA) to determine each UAV's path, with movements structured in epochs. During each epoch, all UAVs can move to an adjacent cell, with the overall movement speed determined by the slowest UAV in the swarm. This approach ensures synchronized operations and avoids desynchronization issues in multi-UAV coordination.

Each gene in a genotype corresponds to a specific UAV's movement at a specific cell during a particular epoch. Before executing the GA, the possible movements from each cell are precomputed based on the presence of adjacent obstacles and map boundaries. This precomputation step reduces the computational overhead during GA execution by limiting the search space to feasible movements only. Movements are encoded as values between 0 and 1, which are mapped to directional intervals (e.g., up, down, left, right) to define the UAV's potential path. The encoded movements are stored in a Movement Map, which serves as the basis for evaluating the UAV's performance and trajectory.

The fitness function, defined in Equation~\eqref{eq:fitness_function_complete}, evaluates the performance of a given genotype \(G\). It returns the epoch count if all cells in the map are visited successfully by that epoch. Otherwise, it penalizes the solution by adding the total number of unvisited cells to the maximum number of epochs allowed, \(\text{MaxEpochs}\). The sets \(C_{\text{visited}}\) and \(C_{\text{total}}\), as defined in Equation~\eqref{eq:cell_definitions}, play a crucial role in this evaluation. \(C_{\text{visited}}\) represents the union of all cells visited by the UAVs, while \(C_{\text{total}}\) is the complete set of cells in the map and \(C_{\text{visit}}(UAV)\) represents all the cells visited by a specific UAV during its traversal of the map. The evaluation of the fitness function follows Algorithm~\ref{alg:Fitness} that evaluates the performance of each candidate solution by simulating the UAVs' movements through the map. The function iterates through epochs, assessing UAV movements, and ensuring that all UAVs contribute to the overall map coverage. If an epoch concludes without any movement or if all cells in the map are covered, the fitness function terminates early, thereby improving computational efficiency. This mechanism allows the GA to prioritize solutions that maximize coverage while minimizing redundant or unnecessary movements.

\begin{equation}
F(G) =
\begin{cases}
\text{epoch}, & \text{if all cells are visited by epoch}, \\
\text{MaxEpochs} + |C_{\text{total}} \setminus C_{\text{visited}}|, & \text{if no UAV moves or MaxEpochs is reached}.
\end{cases}
\label{eq:fitness_function_complete}
\end{equation}

\begin{equation}
\begin{aligned}
C_{\text{visited}} &= \bigcup_{UAV} C_{\text{visit}}(UAV), 
\end{aligned}
\label{eq:cell_definitions}
\end{equation}

\begin{algorithm}[H]
\SetAlgoLined
\label{alg:Fitness}
\KwData{genotype}
\KwResult{Fitness value}
\For {$UAV\leftarrow 1$ \KwTo $NumUAVs$} {
    calculate Movement Map for this UAV from genotype\;
}
initialize each UAV at their initial positions\;
epoch = 1\;
\While {epoch $\leq$ Max num of epochs} {
    \For {$UAV\leftarrow 1$ \KwTo $NumUAVs$} {
        read movement of this UAV from its position in the map\;
        \uIf {movement goes to a cell not visited by this UAV $\And$ movement goes to a cell whithout another  UAV} {
            perform movement
%        } \uElse {
%            return $\emptyset$
        }
    }
    \uIf {no UAV has been moved} {
        return Max num of epochs + Num cells not visited
    } \uElseIf {all of the cells have been visited} {
        return epoch
    }
    epoch $\leftarrow$ epoch + 1\;
}
return Max num of epochs + Num cells not visited
\caption{Fitness function}
\end{algorithm}

\subsection{Limitations}\label{sec:Limitations}

A primary limitation arises from constructing a fixed Movement Map for each UAV, which restricts their ability to revisit cells using different movement paths. This limitation becomes particularly problematic when UAVs encounter dead ends, as observed in Map 5, where some UAVs may halt prematurely due to the lack of alternative movement options. While one possible solution involves evolving additional Movement Maps to provide alternative paths, this study addresses such limitations by leveraging the use of multiple UAVs to achieve complete map coverage. Simulations with a single UAV are included for comparative purposes to highlight the effectiveness of multi-UAV coordination in mitigating these issues.

\section{Results}\label{results}

A significant number of experiments were conducted using the maps described in Section~\ref{sec:Maps}, in conjunction with varying quantities of UAVs and different hyperparameter settings for the Genetic Algorithm (GA). The primary objective of these experiments was to evaluate the system's performance across diverse scenarios, ensuring the robustness and adaptability of the proposed method.

To systematically investigate the effects of these factors, a comprehensive grid search methodology was employed. This approach explored all possible combinations of hyperparameters, allowing for a thorough assessment of the GA's behavior and its ability to optimize UAV Path Planning in different configurations. The experiments aimed to identify optimal settings for each combination of map topology and UAV count, as well as to examine the interaction between these parameters.

The experimental configurations can be categorized into two main groups:
\begin{itemize}
    \item \textbf{Map Configurations:} Variations in map size, obstacle placement.
    \item \textbf{GA Configurations:} Variations in key hyperparameters.
\end{itemize}

Table~\ref{table:summary} summarizes these configurations, providing an overview of the parameter space explored during the grid search. The experimental results offer insights into how the combination of map configurations and GA hyperparameters affects the efficiency, coverage, and computational cost of the proposed system.

\begin{table}[]
    \centering
    \caption{Comprehensive table detailing the values used in each of the 30,000 runs performed.}
    \resizebox{\textwidth}{!}{
    \begin{tabular}{l|c|c|}
        \cline{2-3}
        & \textbf{Variable}                                    & \textbf{Value(s)}                              \\ \hline
        \multicolumn{1}{|l|}{\multirow{2}{*}{\textbf{\begin{tabular}[c]{@{}l@{}}Map \\ Configurations\end{tabular}}}}               & \textbf{Number of UAVs}                              & 1 to 4 UAVs                                 \\ \cline{2-3} 
        \multicolumn{1}{|l|}{}                                                                                                    & \textbf{Number of maps}                              & 6 maps                                      \\ \hline
        \multicolumn{1}{|l|}{\multirow{3}{*}{\textbf{\begin{tabular}[c]{@{}l@{}}Genetic Algorithm\\ Configurations\end{tabular}}}} & \multicolumn{1}{l|}{\textbf{Population sizes}}       & 1000, 2000, 3000, 4000 and 5000 individuals \\ \cline{2-3} 
        \multicolumn{1}{|l|}{}                                                                                                    & \multicolumn{1}{l|}{\textbf{Number of generations}}  & 100, 200, 300, 400, 500 generations         \\ \cline{2-3} 
        \multicolumn{1}{|l|}{}                                                                                                    & \textbf{Parallel runs}                               & 50 runs                                     \\ \hline
    \end{tabular}
    \label{table:summary}
    }
\end{table}

The map configurations consist of the six aforementioned maps, each paired with a varying number of UAVs ranging from 1 to 4. This results in a total of 24 distinct map configurations, covering a diverse set of scenarios in terms of map complexity and swarm size. These configurations were designed to systematically test the algorithm's adaptability to different environmental challenges and UAV swarm dynamics.

On the other hand, the GA configurations were tested with five different population sizes: 1000, 2000, 3000, 4000, and 5000, combined with the number of generations set to 100, 200, 300, 400, and 500. This yields a total of 25 unique GA configurations, enabling a comprehensive evaluation of how these hyperparameters influence the convergence speed, solution quality, and computational efficiency of the algorithm.

Given the stochastic nature of the GA, each combination of map configuration and GA configuration was subjected to 50 parallel runs to ensure statistical reliability and to account for variability in the results. Consequently, a grand total of 30,000 independent runs were conducted in the course of this study.

After completing the experiments, the percentage of successful runs was calculated for each unique configuration defined by a tuple of parameters (map, number of UAVs, population size, maximum number of generations). A successful run was defined as obtaining a solution that fully covered the entire terrain. Table~\ref{tab:PercentCovered} presents the maximum percentage achieved for any GA  configuration, organized by map configuration (rows corresponding to different maps and columns indicating the number of UAVs).

As shown in Table~\ref{tab:PercentCovered}, full terrain coverage was achieved with a single UAV only in the case of the first map, which has no obstacles. For the remaining maps, the presence of corners prevented full coverage with a single UAV, resulting in a maximum percentage of 0\% for these scenarios. Using two UAVs allowed complete coverage of the terrain for maps 1 through 4. For map 2, despite the large number of corners, the UAV starting points being located in two corners enabled full coverage. A similar situation was observed in map 4. However, for map 5, even though the UAVs also started from corners, this was insufficient to achieve full coverage. Regarding map 6, its topology inherently prevents full coverage with three UAVs under the constraints of this study, requiring at least four UAVs.

Deploying three UAVs enabled full coverage of the terrain in map 5. This result can be attributed to the fact that the UAV starting points are located in corners, as mentioned previously. However, for map 6, three UAVs were insufficient to achieve full coverage due to its complex topology, which requires the use of four UAVs for complete terrain coverage.

\begin{table}[]
\centering
\caption{Maximum percentage of configurations in which the UAVs covered the whole map for each map (rows) and each number of UAVs (columns)}
\begin{tabular}{cc|cccc|}
\cline{3-6}
                                           &   & \multicolumn{4}{c|}{Number of UAVs}                                               \\ \cline{3-6} 
                                           &   & \multicolumn{1}{c|}{1}  & \multicolumn{1}{c|}{2}  & \multicolumn{1}{c|}{3}  & 4  \\ \hline
\multicolumn{1}{|c|}{\multirow{6}{*}{Map}} & 1 & \multicolumn{1}{c|}{78\%} & \multicolumn{1}{c|}{100\%} & \multicolumn{1}{c|}{100\%} & 100\% \\ \cline{2-6} 
\multicolumn{1}{|c|}{}                     & 2 & \multicolumn{1}{c|}{0\%} & \multicolumn{1}{c|}{100\%} & \multicolumn{1}{c|}{100\%} & 100\% \\ \cline{2-6} 
\multicolumn{1}{|c|}{}                     & 3 & \multicolumn{1}{c|}{0\%} & \multicolumn{1}{c|}{100\%} & \multicolumn{1}{c|}{100\%} & 100\% \\ \cline{2-6} 
\multicolumn{1}{|c|}{}                     & 4 & \multicolumn{1}{c|}{0\%} & \multicolumn{1}{c|}{100\%} & \multicolumn{1}{c|}{100\%} & 100\% \\ \cline{2-6} 
\multicolumn{1}{|c|}{}                     & 5 & \multicolumn{1}{c|}{0\%} & \multicolumn{1}{c|}{0\%} & \multicolumn{1}{c|}{30\%} & 100\% \\ \cline{2-6} 
\multicolumn{1}{|c|}{}                     & 6 & \multicolumn{1}{c|}{0\%} & \multicolumn{1}{c|}{0\%} & \multicolumn{1}{c|}{0\%} & 100\% \\ \hline
\end{tabular}
\label{tab:PercentCovered}
\end{table}

Tables \ref{tab:PercentCovered_map1_1UAV_2UAV}-\ref{tab:PercentCovered_map6_4UAV} present the percentage of Genetic Algorithm (GA) runs that generated a solution fully covering the terrain, for various combinations of population size and number of generations. A separate table is provided for each map configuration and number of UAVs, excluding cases where all percentages were consistently 0\% or 100\%. For instance, in the case of map 5, at least 3 UAVs are required to cover the entire terrain; thus, the percentages for configurations with 1 or 2 UAVs were consistently 0\%, and these tables are omitted. Similarly, for map 2, configurations with more than 2 UAVs achieved 100\% coverage in all runs, so these tables are not included.

A consistent trend can be observed across the tables, with some isolated exceptions, when comparing configurations that involve similar computational effort. In this context, computational effort is defined as the total number of individuals evaluated, calculated as the product of population size and the number of generations. The observed tendency indicates that larger population sizes combined with fewer generations generally produce better results compared to configurations with smaller population sizes and more generations. For example, in the case of map 4 with 2 UAVs (Table \ref{tab:PercentCovered_map4_2UAV}), the first column of the table shows a higher percentage of successful runs compared to the first row, despite both configurations requiring a similar computational effort.

%\begin{table}
%    \centering
%    \caption{Percentage of executions in which the map was fully covered in each configuration for map 1 and 1 UAV. Rows: population size. Columns: number of generations}
%    \begin{tabular}{@{}c|ccccc@{}} 
%        \hline
%        & 100 & 200 & 300 & 400 & 500 \\
%        \hline
%1000 & 12\% & 34\% & 36\% & 36\% & 20\% \\
%2000 & 12\% & 52\% & 46\% & 64\% & 48\% \\
%3000 & 12\% & 58\% & 54\% & 46\% & 50\% \\
%4000 & 22\% & 46\% & 62\% & 58\% & 54\% \\
%5000 & 26\% & 78\% & 56\% & 72\% & 54\% \\
%        \hline
%    \end{tabular}
%    \label{tab:PercentCovered_map1_1UAV}
%\end{table}

%\begin{table}
%    \centering
%    \caption{Percentage of executions in which the map was fully covered in each configuration for map 1 and 2 UAV. Rows: population size. Columns: number of generations}
%    \begin{tabular}{@{}c|ccccc@{}} 
%        \hline
%        & 100 & 200 & 300 & 400 & 500 \\
%        \hline
%1000 & 94\% & 100\% & 98\% & 98\% & 98\% \\
%2000 & 100\% & 100\% & 100\% & 98\% & 98\% \\
%3000 & 100\% & 100\% & 100\% & 100\% & 100\% \\
%4000 & 100\% & 100\% & 100\% & 100\% & 100\% \\
%5000 & 100\% & 100\% & 100\% & 100\% & 100\% \\
%        \hline
%    \end{tabular}
%    \label{tab:PercentCovered_map1_2UAV}
%\end{table}

\begin{table}
    \centering
    \caption{Percentage of executions in which the map was fully covered in each configuration for map 1 and 1 and 2 UAVs. Rows: population size. Columns: number of generations}
    \resizebox{\textwidth}{!}{\begin{tabular}{@{}c|ccccc|ccccc@{}|}
        \cline{2-11}
        & \multicolumn{5}{|c|}{Generations for 1 UAV} & \multicolumn{5}{|c|}{Generations for 2 UAV}\\
        \hline
\multicolumn{1}{|c|}{Population size} & 100 & 200 & 300 & 400 & 500 & 100 & 200 & 300 & 400 & 500 \\
        \hline
\multicolumn{1}{|c|}{1000} & 12\% & 34\% & 36\% & 36\% & 20\%  & 94\% & 100\% & 98\% & 98\% & 98\% \\
\multicolumn{1}{|c|}{2000} & 12\% & 52\% & 46\% & 64\% & 48\% & 100\% & 100\% & 100\% & 98\% & 98\% \\
\multicolumn{1}{|c|}{3000} & 12\% & 58\% & 54\% & 46\% & 50\%  & 100\% & 100\% & 100\% & 100\% & 100\% \\
\multicolumn{1}{|c|}{4000} & 22\% & 46\% & 62\% & 58\% & 54\% & 100\% & 100\% & 100\% & 100\% & 100\% \\
\multicolumn{1}{|c|}{5000} & 26\% & 78\% & 56\% & 72\% & 54\%  & 100\% & 100\% & 100\% & 100\% & 100\% \\
        \hline
    \end{tabular}}
    \label{tab:PercentCovered_map1_1UAV_2UAV}
\end{table}

\begin{table}
    \centering
    \caption{Percentage of executions in which the map was fully covered in each configuration for map 2 and 2 UAV. Rows: population size. Columns: number of generations}
    \begin{tabular}{@{}c|ccccc@{}|} 
        \cline{2-6}
    & \multicolumn{5}{|c|}{Number of generations} \\
        \hline
\multicolumn{1}{|c|}{Population size} & 100 & 200 & 300 & 400 & 500 \\
        \hline
\multicolumn{1}{|c|}{1000} & 96\% & 100\% & 98\% & 96\% & 100\% \\
\multicolumn{1}{|c|}{2000} & 100\% & 100\% & 100\% & 100\% & 100\% \\
\multicolumn{1}{|c|}{3000} & 100\% & 100\% & 100\% & 100\% & 100\% \\
\multicolumn{1}{|c|}{4000} & 100\% & 100\% & 100\% & 100\% & 100\% \\
\multicolumn{1}{|c|}{5000} & 100\% & 100\% & 100\% & 100\% & 100\% \\
        \hline
    \end{tabular}
    \label{tab:PercentCovered_map2_1UAV}
\end{table}

\begin{table}
    \centering
    \caption{Percentage of executions in which the map was fully covered in each configuration for map 3 and 2 UAV. Rows: population size. Columns: number of generations}
    \begin{tabular}{@{}c|ccccc@{}|} 
        \cline{2-6}
    & \multicolumn{5}{|c|}{Number of generations} \\
        \hline
\multicolumn{1}{|c|}{Population size} & 100 & 200 & 300 & 400 & 500 \\
        \hline
\multicolumn{1}{|c|}{1000} & 100\% & 98\% & 100\% & 98\% & 98\% \\
\multicolumn{1}{|c|}{2000} & 100\% & 100\% & 100\% & 100\% & 100\% \\
\multicolumn{1}{|c|}{3000} & 100\% & 100\% & 100\% & 100\% & 100\% \\
\multicolumn{1}{|c|}{4000} & 100\% & 100\% & 100\% & 100\% & 100\% \\
\multicolumn{1}{|c|}{5000} & 100\% & 100\% & 100\% & 100\% & 100\% \\
        \hline
    \end{tabular}
    \label{tab:PercentCovered_map3_2UAV}
\end{table}

\begin{table}
    \centering
    \caption{Percentage of executions in which the map was fully covered in each configuration for map 4 and 2 UAV. Rows: population size. Columns: number of generations}
    \begin{tabular}{@{}c|ccccc@{}|} 
        \cline{2-6}
    & \multicolumn{5}{|c|}{Number of generations} \\
        \hline
\multicolumn{1}{|c|}{Population size} & 100 & 200 & 300 & 400 & 500 \\
        \hline
\multicolumn{1}{|c|}{1000} & 76\% & 96\% & 84\% & 92\% & 96\% \\
\multicolumn{1}{|c|}{2000} & 98\% & 98\% & 100\% & 98\% & 100\% \\
\multicolumn{1}{|c|}{3000} & 98\% & 100\% & 100\% & 100\% & 100\% \\
\multicolumn{1}{|c|}{4000} & 100\% & 100\% & 100\% & 100\% & 100\% \\
\multicolumn{1}{|c|}{5000} & 100\% & 100\% & 100\% & 100\% & 100\% \\
        \hline
    \end{tabular}
    \label{tab:PercentCovered_map4_2UAV}
\end{table}

%\begin{table}
%    \centering
%    \caption{Percentage of executions in which the map was fully covered in each configuration for map 5 and 3 UAV. Rows: population size. Columns: number of generations}
%    \begin{tabular}{@{}c|ccccc@{}} 
%        \hline
%        & 100 & 200 & 300 & 400 & 500 \\
%        \hline
%1000 & 2\% & 4\% & 10\% & 10\% & 6\% \\
%2000 & 4\% & 6\% & 12\% & 14\% & 10\% \\
%3000 & 14\% & 20\% & 12\% & 20\% & 16\% \\
%4000 & 14\% & 12\% & 16\% & 18\% & 30\% \\
%5000 & 18\% & 18\% & 18\% & 22\% & 30\% \\
%        \hline
%    \end{tabular}
%    \label{tab:PercentCovered_map5_3UAV}
%\end{table}

%\begin{table}
%    \centering
%    \caption{Percentage of executions in which the map was fully covered in each configuration for map 5 and 4 UAV. Rows: population size. Columns: number of generations}
%    \begin{tabular}{@{}c|ccccc@{}} 
%        \hline
%        & 100 & 200 & 300 & 400 & 500 \\
%        \hline
%1000 & 74\% & 90\% & 96\% & 90\% & 88\% \\
%2000 & 88\% & 98\% & 98\% & 100\% & 96\% \\
%3000 & 96\% & 100\% & 100\% & 98\% & 98\% \\
%4000 & 94\% & 100\% & 100\% & 100\% & 96\% \\
%5000 & 98\% & 100\% & 100\% & 100\% & 100\% \\
%    \hline
%    \end{tabular}
%    \label{tab:PercentCovered_map5_4UAV}
%\end{table}

\begin{table}
    \centering
    \caption{Percentage of executions in which the map was fully covered in each configuration for map 5 and 3 and 4 UAVs. Rows: population size. Columns: number of generations}
    \resizebox{\textwidth}{!}{\begin{tabular}{@{}c|ccccc|ccccc@{}|} 
        \cline{2-11}
        & \multicolumn{5}{|c|}{Generations for 3 UAV} & \multicolumn{5}{|c|}{Generations for 4 UAV}\\
        \hline
\multicolumn{1}{|c|}{Population size} & 100 & 200 & 300 & 400 & 500 & 100 & 200 & 300 & 400 & 500 \\
        \hline
\multicolumn{1}{|c|}{1000} & 2\% & 4\% & 10\% & 10\% & 6\%  & 74\% & 90\% & 96\% & 90\% & 88\% \\
\multicolumn{1}{|c|}{2000} & 4\% & 6\% & 12\% & 14\% & 10\%  & 88\% & 98\% & 98\% & 100\% & 96\% \\
\multicolumn{1}{|c|}{3000} & 14\% & 20\% & 12\% & 20\% & 16\%  & 96\% & 100\% & 100\% & 98\% & 98\% \\
\multicolumn{1}{|c|}{4000} & 14\% & 12\% & 16\% & 18\% & 30\%  & 94\% & 100\% & 100\% & 100\% & 96\% \\
\multicolumn{1}{|c|}{5000} & 18\% & 18\% & 18\% & 22\% & 30\%  & 98\% & 100\% & 100\% & 100\% & 100\% \\
        \hline
    \end{tabular}}
    \label{tab:PercentCovered_map5_3UAV_4UAV}
\end{table}

\begin{table}
    \centering
    \caption{Percentage of executions in which the map was fully covered in each configuration for map 6 and 4 UAV. Rows: population size. Columns: number of generations}
    \begin{tabular}{c@{}c|ccccc@{}|} 
        \cline{3-7}
\multicolumn{1}{c}{} & & \multicolumn{5}{c|}{Number of UAVs}\\
        \cline{3-7}
       \multicolumn{1}{c}{} & & 100 & 200 & 300 & 400 & 500 \\
        \hline
\multicolumn{1}{|c|}{\multirow{6}{*}{Population size}} & \multicolumn{1}{|c|}{1000} & 52\% & 64\% & 86\% & 88\% & 82\% \\
\multicolumn{1}{|c|}{} & \multicolumn{1}{|c|}{2000} & 96\% & 90\% & 98\% & 92\% & 94\% \\
\multicolumn{1}{|c|}{} & \multicolumn{1}{|c|}{3000} & 96\% & 100\% & 100\% & 98\% & 100\% \\
\multicolumn{1}{|c|}{} & \multicolumn{1}{|c|}{4000} & 98\% & 98\% & 100\% & 100\% & 100\% \\
\multicolumn{1}{|c|}{} & \multicolumn{1}{|c|}{5000} & 100\% & 100\% & 100\% & 100\% & 100\% \\
        \hline
    \end{tabular}
    \label{tab:PercentCovered_map6_4UAV}
\end{table}

In addition to achieving complete terrain coverage, another objective of this study is to minimize the number of movements required to do so. For each map configuration (defined by the map and the number of UAVs), the Genetic Algorithm (GA) configuration (population size and number of generations) that, on average, required the fewest epochs to cover the entire terrain was identified. This was achieved by analyzing the best solution from each run for each GA configuration and averaging the number of epochs needed by these solutions. It is important to note that only the epochs from runs that successfully covered the entire terrain were included in this average. Consequently, for each map configuration, one average value is associated with each GA configuration. This allows the selection of the GA configuration that corresponds to the lowest average number of epochs for each map configuration.

Table~\ref{tab:BestConfigurations} presents three values for each map configuration. The top value corresponds to the average number of epochs required by the solutions. The middle and lower values represent the GA configuration (population size and number of generations) in which this average was achieved. By comparing these values with those in Table~\ref{tab:OptimalNumMovements}, it can be observed that the number of movements often equals the optimal value. For maps 1 and 3, the optimal value is reached in most runs that successfully covered the terrain. This is attributed to the topology of these maps, which allows them to be covered without revisiting cells. In these cases, the GA successfully identifies these optimal solutions.

For maps 2 and 4, the topology does not permit complete traversal without revisiting some cells, making it impossible to achieve the values in Table~\ref{tab:OptimalNumMovements}. However, the lowest number of epochs required by any of the 1250 executions performed for each map configuration has been considered the best possible value in this study. These values are shown in Table~\ref{tab:MinEpochs}. While this method does not theoretically guarantee optimality, the extensive number of experiments performed suggests that these values are highly likely to be optimal.

A comparison of the values in Table~\ref{tab:MinEpochs} with the average epochs for solutions that covered the entire terrain (Table~\ref{tab:BestConfigurations}) reveals that, in most cases, the number of epochs is equal to or very close to the minimum. The only exception occurs in map 6, where the average value is significantly higher. This discrepancy arises because the optimal path for UAVs in map 6 is nearly unique. Consequently, partial solutions that cover the entire terrain are unlikely to converge to this global optimal solution. This behavior is not observed in the other maps, where solutions that cover the terrain but are not optimal tend to be closer to the optimal solution in the search space.

\begin{table}
    \centering
    \caption{Best configurations for each map and number of UAV (average number of epochs / population size / number of generations).}
    \begin{tabular}{|c|c|c|c|c|}
\hline
\textbf{Map} & \textbf{1 UAV} & \textbf{2 UAVs} & \textbf{3 UAVs} & \textbf{4 UAVs} \\ \hline
1 & 48.0 / 1000 / 100 & 24.0 / 4000 / 300 & 16.32 / 4000 / 500 & 12.04 / 4000 / 500 \\ \hline
2 & - & 13.88 / 5000 / 200 & 9.0 / 2000 / 100 & 7.0 / 1000 / 100 \\ \hline
3 & - & 13.0 / 2000 / 100 & 9.0 / 1000 / 100 & 6.0 / 1000 / 100 \\ \hline
4 & - & 23.0 / 5000 / 500 & 16.04 / 3000 / 400 & 11.0 / 2000 / 400 \\ \hline
5 & - & - & 25.0 / 1000 / 100 & 16.1 / 4000 / 400 \\ \hline
6 & - & - & - & 21.49 / 1000 / 500 \\ \hline
\end{tabular}
    \label{tab:BestConfigurations}
\end{table}

\begin{table}[]
\centering
\caption{Minimum number of epochs needed for each map configuration}
\begin{tabular}{cc|cccc|}
\cline{3-6}
                                           &   & \multicolumn{4}{c|}{Number of Epochs}                                               \\ \cline{3-6} 
                                           &   & \multicolumn{1}{c|}{1}  & \multicolumn{1}{c|}{2}  & \multicolumn{1}{c|}{3}  & 4  \\ \hline
\multicolumn{1}{|c|}{\multirow{6}{*}{Map}} & 1 & \multicolumn{1}{c|}{48} & \multicolumn{1}{c|}{24} & \multicolumn{1}{c|}{16} & 12 \\ \cline{2-6} 
\multicolumn{1}{|c|}{}                     & 2 & \multicolumn{1}{c|}{43} & \multicolumn{1}{c|}{13} & \multicolumn{1}{c|}{9} & 7 \\ \cline{2-6} 
\multicolumn{1}{|c|}{}                     & 3 & \multicolumn{1}{c|}{58} & \multicolumn{1}{c|}{13} & \multicolumn{1}{c|}{9} & 6 \\ \cline{2-6} 
\multicolumn{1}{|c|}{}                     & 4 & \multicolumn{1}{c|}{79} & \multicolumn{1}{c|}{23} & \multicolumn{1}{c|}{15} & 11 \\ \cline{2-6} 
\multicolumn{1}{|c|}{}                     & 5 & \multicolumn{1}{c|}{109} & \multicolumn{1}{c|}{51} & \multicolumn{1}{c|}{23} & 15 \\ \cline{2-6} 
\multicolumn{1}{|c|}{}                     & 6 & \multicolumn{1}{c|}{128} & \multicolumn{1}{c|}{64} & \multicolumn{1}{c|}{41} & 15 \\ \hline
\end{tabular}
\label{tab:MinEpochs}
\end{table}

Table~\ref{tab:TrainingTimes} presents the average training times for each map and Genetic Algorithm (GA) configuration. As in previous analyses, the times were calculated exclusively from runs that successfully covered the entire map. 

The results in Table~\ref{tab:TrainingTimes} indicate that training times were relatively low, with the maximum observed time slightly exceeding 10 minutes. All experiments were conducted on Intel Xeon Ice Lake 8352 nodes without the use of GPUs, highlighting the computational efficiency of the proposed approach under these hardware conditions.

\begin{table}
   \centering
\caption{Range (minimum / maximum) of training times (seconds) by number of UAVs on each map.}
\label{tab:training_times}
\begin{tabular}{|c|c|c|c|c|}
\hline
    \textbf{Map} & \textbf{1 UAV} & \textbf{2 UAVs} & \textbf{3 UAVs} & \textbf{4 UAVs} \\ \hline
1 & 17.2 / 625.08 & 15.88 / 511.14 & 11.52 / 644.99 & 10.40 / 626.93 \\ \hline
2 & 12.87 / 497.48 & 12.46 / 377.44 & 12.36 / 583.84 & 10.57 / 659.91 \\ \hline
3 & 16.59 / 619.09 & 15.25 / 617.79 & 15.18 / 653.08 & 10.33 / 644.55 \\ \hline
4 & 18.01 / 687.44 & 15.11 / 679.44 & 14.87 / 651.99 & 12.98 / 639.54 \\ \hline
5 & 19.13 / 395.45 & 18.25 / 362.15 & 14.92 / 403.84 & 11.79 / 659.91 \\ \hline
6 & 11.52 / 528.21 & 10.41 / 532.15 & 10.11 / 555.12 & 10.05 / 659.97 \\ \hline
\end{tabular}
    \medskip
    \label{tab:TrainingTimes}
\end{table}

\begin{figure}[!ht]
        \centering
        \subfigure[UAV 1]{
        \includegraphics[scale=0.4]{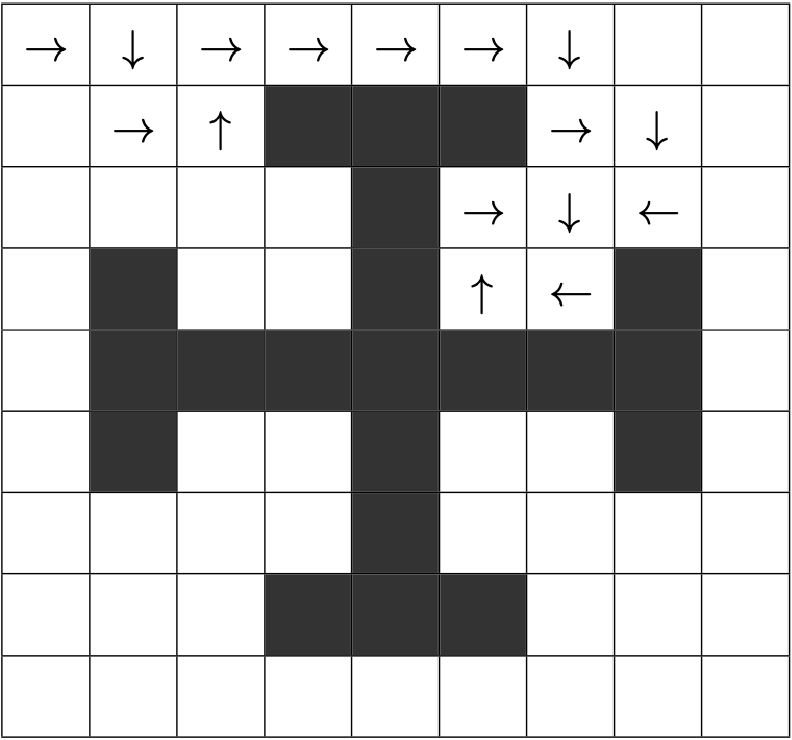}
        \label{fig:movement_set1}
        }
    	\subfigure[UAV 2]{\includegraphics[scale=0.4]{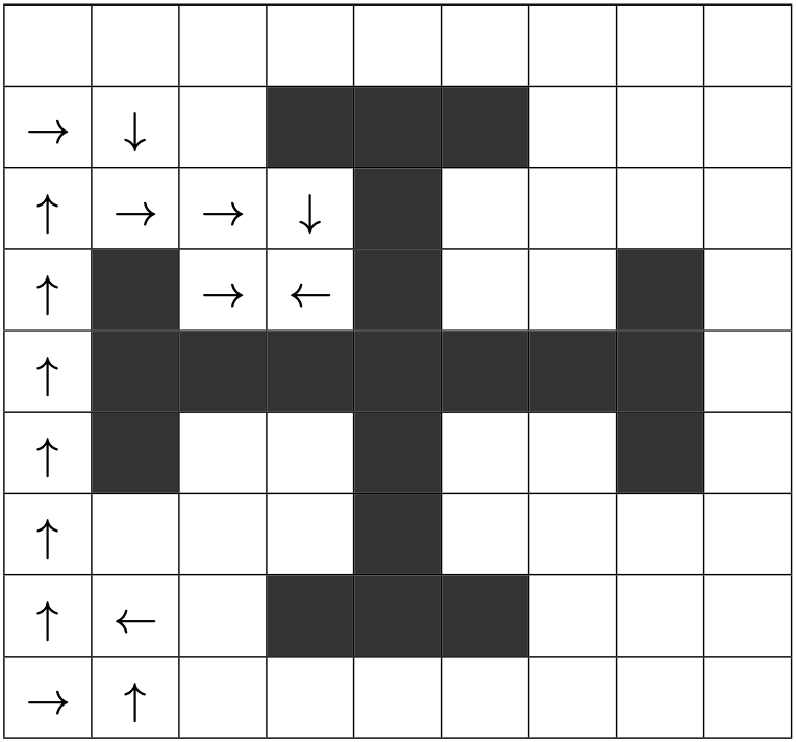}
        \label{fig:movement_set2}}
    	\subfigure[UAV 3]{\includegraphics[scale=0.38]{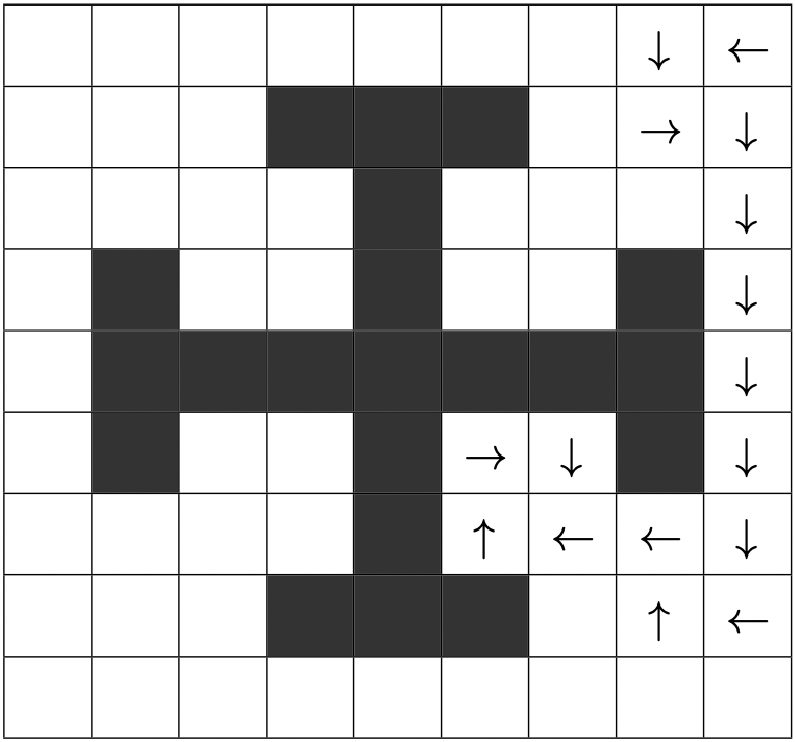}
        \label{fig:movement_set3}}
    	\subfigure[UAV 4]{\includegraphics[scale=0.38]{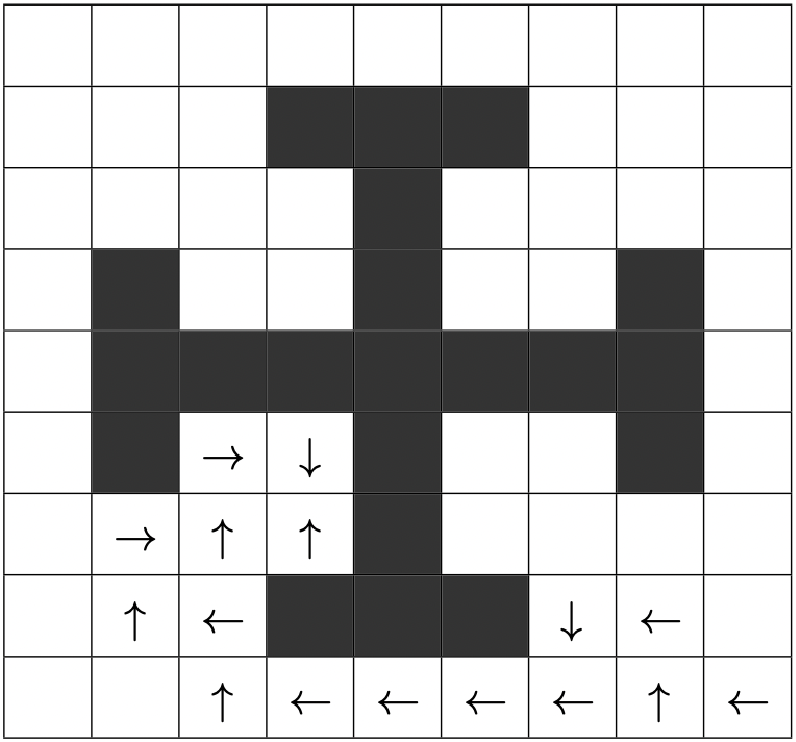}
        \label{fig:movement_set4}}
        \caption{Paths obtained for each UAV on map 6.  Paths are shown differentiated for each UAV for easier visualization. In fact, these paths are executed simultaneously..}
    \label{fig:movement_set}
\end{figure}

Figure~\ref{fig:movement_set} illustrates the optimal path identified for map 6 using 4 UAVs. This map was selected due to its complexity and the challenges it presents in achieving the optimal path, as discussed previously. 

The Figure demonstrates that, to fully cover the terrain in this map, each UAV must traverse a cell that has already been visited by another UAV. These critical cells are located on the diagonal relative to the starting position of each UAV. During the initial stages of movement, each UAV is required to visit its respective diagonal cell, even though this cell will later be traversed by another UAV. Failure to include these movements prevents complete coverage of the map.

Table~\ref{tab:results_grouped} compares the results of the proposed GA-based method with those obtained in a prior RL-based study \citep{puente2024q}, which utilized a Reinforcement Learning (RL) approach. The table presents the results for different map configurations and numbers of UAVs, reported as the mean and standard deviation of the epochs required to cover the entire map. This comparison aims to highlight the differences in performance between the two methods when tested on identical scenarios.

The results show a clear advantage of the proposed GA-based method over the RL-based approach across all maps and UAV configurations. Specifically, the proposed method consistently requires significantly fewer epochs to achieve complete coverage. For instance, in Map 1 with a single UAV, the GA method required an average of $124.60 \pm 2.42$ movements, while the RL-based method needed $17297.80 \pm 2186.93$ movements. Similar trends are observed as the number of UAVs increases, with the GA method demonstrating lower mean values and smaller standard deviations, indicating greater efficiency and stability.

Maps with more complex topologies, such as Maps 5 and 6, exhibit a similar pattern. For Map 6, the GA method required $161.68 \pm 2.08$ movements with a single UAV, compared to $10764.40 \pm 907.76$ movements for the RL method. Even as the number of UAVs increases, the GA method maintains its superior performance, as seen with 4 UAVs, where the GA method required $67.76 \pm 2.06$ movements, significantly fewer than the $9392.40 \pm 2002.47$ movements required by the RL method.

Overall, the results highlight the robustness and efficiency of the GA-based approach, especially in scenarios with complex terrain and varying UAV configurations. The smaller standard deviations observed for the GA method indicate that it delivers more consistent performance compared to the RL method, which exhibits high variability in its results. This consistency is particularly important in applications where reliable performance is critical.

\begin{table}[h!]
\centering
\caption{Comparison of RL and GA results for different UAVs and map sizes. Results are shown as mean ± standard deviation.}
\label{tab:results_grouped}
\begin{tabular}{|c|c|c|c|}
\hline
\textbf{Map} & \textbf{UAVs} & \textbf{Proposed Method} & \textbf{\cite{puente2024q}} \\
\hline
\multirow{4}{*}{\textbf{1}} & 1 & $124.60 \pm 2.42$ & $17297.80 \pm 2186.93$ \\
                                                & 2 & $72.94 \pm 2.15$ & $9117.80 \pm 7924.43$ \\
                                                & 3 & $54.94 \pm 2.12$ & $6032.80 \pm 5877.52$ \\
                                                & 4 & $45.78 \pm 2.18$ & $6713.20 \pm 6773.66$ \\
\hline
\multirow{4}{*}{\textbf{2}} & 1 & $46.88 \pm 1.38$ & $15086.00 \pm 3910.30$ \\
                              & 2 & $23.88 \pm 1.02$ & $1265.00 \pm 891.23$ \\
                              & 3 & $16.00 \pm 0.76$ & $571.40 \pm 374.07$ \\
                              & 4 & $13.14 \pm 0.86$ & $265.60 \pm 137.48$ \\
\hline
\multirow{4}{*}{\textbf{3}} & 1 & $67.18 \pm 0.98$ & $22562.60 \pm 3366.92$ \\
                              & 2 & $38.76 \pm 1.19$ & $2619.20 \pm 1780.14$ \\
                              & 3 & $28.96 \pm 1.43$ & $5172.80 \pm 8840.37$ \\
                              & 4 & $25.00 \pm 1.28$ & $5128.00 \pm 7363.21$ \\
\hline
\multirow{4}{*}{\textbf{4}} & 1 & $97.90 \pm 2.43$ & $16022.80 \pm 1452.18$ \\
                              & 2 & $56.96 \pm 2.03$ & $13107.80 \pm 6544.31$ \\
                              & 3 & $42.62 \pm 2.07$ & $8800.00 \pm 7003.98$ \\
                              & 4 & $36.10 \pm 1.56$ & $4188.80 \pm 2619.40$ \\
\hline
\multirow{4}{*}{\textbf{5}} & 1 & $134.58 \pm 1.82$ & $13657.80 \pm 1813.33$ \\
                              & 2 & $81.36 \pm 2.18$ & $13030.60 \pm 1048.04$ \\
                              & 3 & $63.04 \pm 2.16$ & $10396.00 \pm 2789.60$ \\
                              & 4 & $54.66 \pm 1.97$ & $10035.80 \pm 4118.25$ \\
\hline
\multirow{4}{*}{\textbf{6}} & 1 & $161.68 \pm 2.08$ & $10764.40 \pm 907.76$ \\
                              & 2 & $98.94 \pm 1.98$ & $9130.00 \pm 1379.18$ \\
                              & 3 & $77.46 \pm 2.32$ & $7759.60 \pm 1946.93$ \\
                              & 4 & $67.76 \pm 2.06$ & $9392.40 \pm 2002.47$ \\
\hline
\end{tabular}
\end{table}

\section{Conclusions}\label{conclusions}

This study has demonstrated the effectiveness of Genetic Algorithms (GAs) in determining UAV paths with the primary objective of achieving complete terrain coverage. Additionally, a secondary objective of minimizing flight time, quantified as the number of UAV movements required, was also addressed. To this end, six distinct maps with varying levels of complexity were designed, a tailored Genetic Algorithm was developed, and an extensive set of experiments was conducted.

The experimental results highlight the robustness and versatility of the proposed GA-based approach. For each map configuration (defined by the map topology and the number of UAVs), a large number of GA configurations were evaluated, and numerous configurations successfully achieved full terrain coverage. Furthermore, the secondary objective of minimizing flight time was consistently met in most scenarios. By calculating the theoretical lower bound of the optimal number of movements prior to the experiments, it was observed that the proposed method reached this optimal value in many cases. For simpler maps, such as Maps 1 and 3, the optimal number of movements was achieved in nearly all successful runs due to their straightforward topology, which allowed full coverage without revisiting cells.

For more complex maps, where revisiting cells is necessary due to their topology, the method achieved solutions that closely approached the optimal value, demonstrating its efficiency even under challenging conditions. For the most complex maps, the results show that the GA method consistently outperformed the Reinforcement Learning (RL)-based approach developed with the same maps. Specifically, the GA method required significantly fewer epochs and exhibited greater stability, as reflected in the smaller standard deviations of its results. This highlights the reliability of the GA-based method in scenarios with intricate terrain and varying UAV configurations.

Another key strength of the proposed approach is its computational efficiency. Training times across all map configurations and GA parameters were remarkably low, with the maximum training time slightly exceeding 10 minutes, even without the use of GPUs. This efficiency underscores the adaptability of the system, enabling it to respond dynamically to changes in the environment or operational requirements.

In summary, this paper presents a novel method for UAV swarm Path Planning based on Genetic Algorithms. The method excels in achieving complete terrain coverage, minimizing flight time, and maintaining high computational efficiency. These characteristics make it a promising tool for a wide range of applications that require UAV swarm coordination, including surveillance, search and rescue, agricultural monitoring, and environmental mapping. Its demonstrated efficiency and stability suggest that it could be deployed effectively in both real-time and pre-planned operations, offering significant potential for future research and practical implementations.

\section{Future Work}\label{future_work}

As explained several times in the paper, this system has a limitation that has been deliberately left in order to study the exploration of complex maps. As a first future work that can be derived from this, a similar system could be developed that would allow a UAV to visit cells that have been previously traversed. To this end, one possibility would be to make the genotype of an individual AG contain not just one movement map for each UAV, but more than one. In this way, each UAV would have a primary movement map that it would use in the cells it has not visited, and a secondary one that it would use in the cells already visited by itself.

In addition, in this work all UAVs have been synchronized in such a way that the movements of all UAVs occur at the same time. If the UAVs have different speeds, this means that when each one leaves its cell to reach the next one, the faster ones will have to wait for the slower ones. One possible future work could be to take advantage of the different speeds to further decrease the time required to traverse the entire map.

Another possibility for future work would be to contemplate a more complex task for the UAVs, where they would have to traverse the entire map, and then return to the starting point. To do this, it would again be necessary to consider the possibility of using several movement maps for each UAV.

\section*{Funding}

This project was supported by the General Directorate of Culture, Education, and University Management of Xunta de Galicia [grant number ED431D 2017/16]. This work was also funded by the grant for the consolidation and structuring of competitive research units [grant number ED431C 2022/46] from the General Directorate of Culture, Education and University Management of Xunta de Galicia, and the CYTED network [grant number PCI2018\_093284] funded by the Spanish Ministry of Innovation and Science. This project was also supported by the General Directorate of Culture, Education and University Management of Xunta de Galicia PRACTICUM DIRECT [grant number IN845D-2020/03].





%\begin{thebibliography}{00}
\bibliographystyle{elsarticle-harv} 
\bibliography{references.bib}

%\end{thebibliography}
\end{document}